\documentclass{article}

\usepackage[final]{corl_2020} 

\usepackage{url}
\usepackage{amssymb}
\usepackage{amsmath}
\usepackage{cite}
\usepackage{comment}
\usepackage{color}
\usepackage{alltt}
\usepackage{cleveref}
\usepackage{tikz}
\usepackage{pgfplots}
\usepackage{pgf-umlsd}
\usepackage{multirow}
\usepackage{algorithm}
\usepackage[noend]{algpseudocode}
\usepackage{siunitx}
\usepackage{paralist}
\usepackage{tabularx}
\usepackage{ifthen}
\usepackage{threeparttable}
\usepackage{nth}
\usepackage{caption}
\usepackage[keeplastbox]{flushend}
\usepackage{bm}
\usepackage{scalefnt}
\newcounter{num}

\newcommand{\rnum}[1]{\setcounter{num}{#1} \roman{num}}

\makeatletter
\newcommand{\figcaption}[1]{\def\@captype{figure}\caption{#1}}
\newcommand{\tblcaption}[1]{\def\@captype{table}\caption{#1}}
\makeatother

\title{Collision-free Path Planning \\ in the Latent Space through cGANs \\}
\author{
  Tomoki Ando\\
  Waseda Univ. \& AIST, Japan\\
  \texttt{tomoki\_a@fuji.waseda.jp} \\
  \And
  Hiroki Mori\\
  Waseda Univ. \& AIST, Japan \\
  Univ. of Cergy-Pontoise, France\\
  \texttt{mori@idr.ias.sci.waseda.ac.jp} \\
  \And
  Ryota Torishima\\
  Waseda Univ. \& AIST, Japan\\
  \texttt{r.torishima@fuji.waseda.jp} \\
  \And
  Kuniyuki Takahashi\\
  Preferred Networks, Japan\\
  \texttt{takahashi@preferred.jp} \\
  \And
  Shoichiro Yamaguchi\\
  Preferred Networks, Japan\\
  \texttt{guguchi@preferred.jp} \\
  \And
  Daisuke Okanohara\\
  Preferred Networks, Japan\\
  \texttt{hillbig@preferred.jp} \\
  \And 
  Tetsuya Ogata\\
  Waseda Univ. \& AIST, Japan\\
  \texttt{ogata@waseda.jp} \\
}

\begin{document}
\maketitle
\begin{abstract} 
We show a new method for collision-free path planning by cGANs by mapping its latent space to only the collision-free areas of the robot joint space.
Our method simply provides this collision-free latent space after which \textbf{any} planner, using \textbf{any} optimization conditions, can be used to generate the most suitable paths on the fly.
We successfully verified this method with a simulated two-link robot arm.
\end{abstract}

\section{Introduction}
\label{sec:introduction}
Collision-free path planning is essential to ensure safety and to prevent the robot from harming itself and its surroundings.
There are three important factors in robot path planning: \rnum{1}~)~\textbf{\emph{Adaptation}}, \rnum{2}~)~\textbf{\emph{Customizability}}, and \rnum{3}~)~\textbf{\emph{Scalability of computation}}.
\rnum{1}~)~Robots need to adapt quickly to a new situation, which requires appropriate path planning for the placement of unknown obstacles.
\rnum{2}~)~Robots need to be customizable for different situations.
There are multiple (infinite) paths from a given start to a goal, and it is necessary to choose the optimal path depending on the environment and situation.
In the real environment, we want to perform path planning not only to avoid collision-obstacles, but also to satisfy other criteria such as the efficiency of the robot`s movements and the speed of its movements.
\rnum{3}~)~These path planning operations should be calculable,  even when there is a large number of obstacles, since it generally takes a long time to collision-check for obstacles.
In other words, calculation time should scale well with the number of obstacles.

\begin{figure}[tb]
    \centering
    \includegraphics[width=0.40\linewidth]{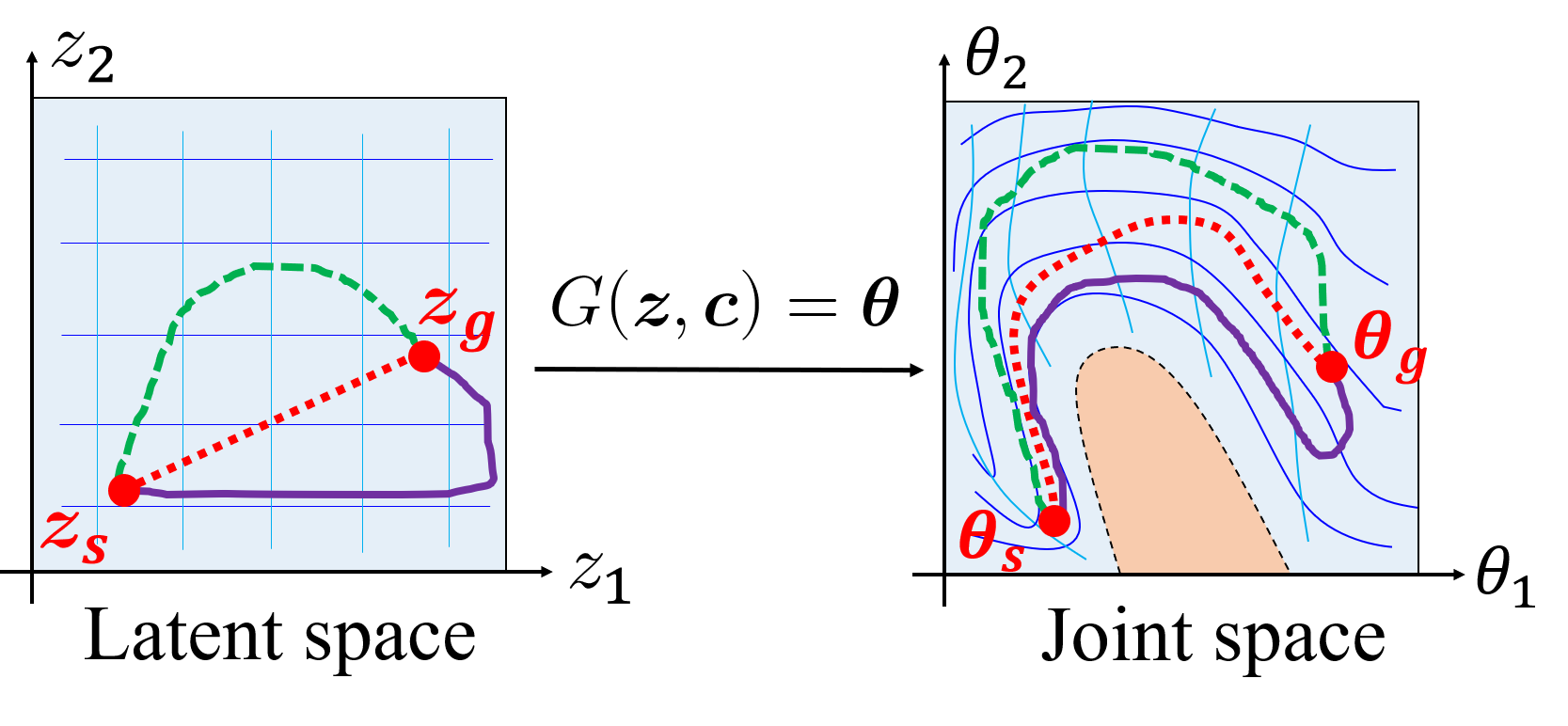}
    \caption{
		Collision-free path planning for robot arm using latent space of cGANs.
		Latent variables $\boldsymbol{z}_{s}$ and $\boldsymbol{z}_{g}$ that correspond to the start $\boldsymbol{\theta}_{s}$ and goal $\boldsymbol{\theta}_{g}$ postures of the robot arm.
    	Any path in the latent space $\boldsymbol{z}_{s}:\boldsymbol{z}_{g}$ is mapped to the collision-free path in the joint space $\boldsymbol{\theta}_{s}:\boldsymbol{\theta}_{g}$ by using Generator ${G}$ with condition $\boldsymbol{c}$ as obstacles information.}
    \label{fig:method_transform}
    \vspace{-4mm}
\end{figure}

It is challenging to meet these three requirements in existing methods (see \Cref{sec:related works}).
Contrary to traditional planning in Cartesian or joint space, we propose to plan within a new collision-free space.
This method uses Conditional Generative Adversarial Networks (cGANs) to map its latent space to only the collision-free areas of the robot joint space, such that the robot does not collide with obstacles if a path is planned within this latent space (See Figure~\ref{fig:method_transform}).
That is, selecting any point in the latent space yields a certain robot pose that does not collide with obstacles.
The mappings from the latent space to joint space \textbf{\emph{adapt}} to obstacles that are given to the cGANs as conditions.
GANs are used because it is a powerful generative model and its latent space can be freely determined in advance; GANs are also highly customizable.
Because the latent space is collision-free and any point on a line connecting any two points is also within this latent space (since this is a convex space), a collision-free path can be generated by connecting the start and goal points with any arbitrary line or curve in the latent space within a domain of definition.
Then, the joint trajectory corresponding to the trajectory planned in latent space is acquired through the learned mappings.
Since we separated the learning of the mappings and the actual path planning (or trajectory including optimizations), we can generate \textbf{any} trajectory we want on the fly, for \textbf{any} optimization parameters that we want without considering collisions, making our method highly \textbf{\emph{customizable}}.
Furthermore, the computation cost does not depend on the number of obstacles because it does not require collision detection, making its computation \textbf{\emph{scalable}}.
We verify this method on a simulated 2D robot arm.

\section{Related Works}
\label{sec:related works}
There are mainly two types of path planning methods: model-based and learning-based methods.
The following two model-based methods are the most common: Design functions for obstacles and goals (e.g., potential fields~\citep{warren1990multiple, li2012efficient}), search and optimization (e.g. RRT~\citep{lavalle1998rapidly, lavalle2001randomized} and $A^{\ast}$~\citep{hart1968formal}).
Methods which are a combination of these are also proposed and generally show improved results~\citep{naderi2015rtrrt, ahmed2015intelligent,ahmed2017potential, tahir2018potentially}.
While model-based methods can reliably avoid obstacles, their adaptability to various environments in real-time is limited since these methods require specific function design and adjustment of parameters for each situation in advance, not to mention the huge computational searching cost.
In addition, sometimes certain conditions need to be optimized depending on the purpose, such as the shortest traveling distance in end-effector space or joint space~\citep{Lalibertk1994RedundantManipulator}
or minimum jerk change~\citep{Flash1985MJ}; usually multiple or infinite paths for the same start and goal states exist, each of them optimized for different purposes.
As model-based methods are usually calculated according to certain conditions/criteria in advance, other calculations need to be performed when these criteria change.
In other words, \textbf{model-based methods lack \emph{scalability} and \emph{customizability}}.
The data collected by the model-based methods can be used to train learning-based algorithms, particularly deep learning~\citep{srinivas2018universal, aviv2016value, ichter2018learning, angelina2019learning, kumar2019lego, Terasawa20203d, ota2020efficient}.
These algorithms can infer a path for a new environment in a short time if it has trained sufficiently in advance.
However, learning-based methods have the challenge that only one or a few paths can be generated, and what kind of paths are generated depends on the training data.
For example, if naive RRT is used as training data, only collision-free paths to the goal will be generated during inference, usually without taking any additional constraints into account that naive RRT also does not.
Usually, \textbf{learning-based methods lack \emph{customizability}}.

In~\citep{kutsuzawa2019motion, ahmed2018motion}, the authors studied the generation of multiple trajectories.
Since the target of~\citep{kutsuzawa2019motion} was to generate various trajectories in environments with no obstacles, obstacle avoidance was out of their scope.
Our proposed method is to plan paths in a \textbf{collision-free space} which are mapped from the latent space to joint space.
Since the trajectory of~\citep{ahmed2018motion} is fixed once it is generated, at best, only the optimal trajectory among the ones generated can be selected, which is not necessarily the best for the situation at hand.
Thus, they have to generate trajectories until one of them satisfies the criteria necessary for the situation, but they are generated randomly and the method does not provide a way to define optimality.
To address this issue, our method does not directly output the trajectories, but simply provides a collision-free space after which any planner, using any optimization conditions, can be used to generate \textbf{the most suitable paths}.

The contribution of this research is to realize optimized path planning with the three important factors; \rnum{1}~)~\textbf{\emph{Adaptation}}, \rnum{2}~)~\textbf{\emph{Customizability}}, and \rnum{3}~)~\textbf{\emph{Scalability of computation}}.
\section{Proposed Method}
\label{sec:method}
We propose a method that maps the latent space of cGANs to the collision-free area of the robot joint space, so that the robot learns not to collide with obstacles.
Thus, any planned path in that latent space can be associated with a collision-free path in joint space.
The mapping from the latent space to joint space \textbf{\emph{adapts}} accordingly to the obstacles given to cGANs as conditions.
Unlike path planners in joint space, since the mapping and path planning phases are separated, any path planner can be used in the trained latent space (where any point is collision-free) without taking obstacles into account since there simply are none in the latent space, making our method highly \textbf{\emph{customizable}}.
The computational cost is also lower since collision check calculations are no longer necessary, making our method also \textbf{\emph{scalable}}.
The correspondence from the latent space to joint space is trained by cGANs, which uses a min-max game between a Generator ($G$) and a Discriminator ($D$), and both are optimized alternately with the following objective function:
\begin{equation}
    \label{eq:loss_all}
    \min_{G} \max_{D} V \left(D, G\right) =
        \mathcal{L}_{\mathrm{GAN}} \left(D, G\right) +
        \mathcal{L}_{\mathrm{collision}} \left(D\right) +
        \mathcal{L}_{\mathrm{identity}} \left(G\right)
\end{equation} 
\begin{enumerate}
    \item $\mathcal{L}_{\mathrm{GAN}}$: The main loss function to learn the mapping from the latent space to joint space.
    \item $\mathcal{L}_{\mathrm{collision}}$: Loss function to learn various situations with obstacle, with less collision than non-collision data.
    \item $\mathcal{L}_{\mathrm{identity}}$: Loss function to learn from the latent space to joint space continuously without ``twist'' or ``distortion''.
    In addition, for the purpose of path planning using a robot arm, rapid or sudden changes in movement are discouraged.
\end{enumerate}

\begin{figure}[tb]
    \centering
     \includegraphics[width=0.95\textwidth]{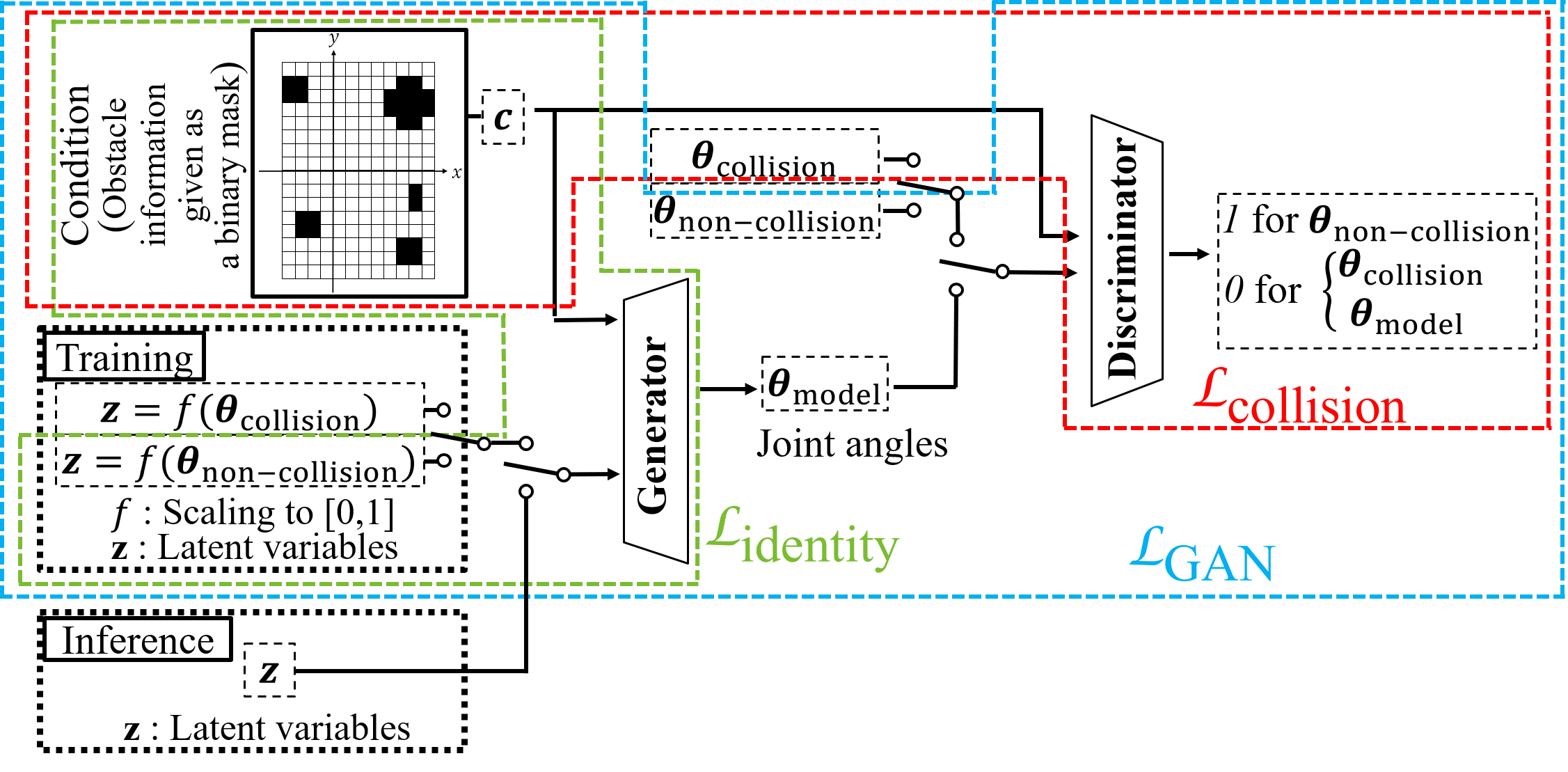}
     \caption{Structure of collision-free path planning model using cGANs.
     }
     \label{fig:network}
\end{figure}

\subsection{$\mathcal{L}_{\mathrm{GAN}}$: Acquisition of Latent Expression}
To acquire the correspondence from the latent space to joint space, cGANs are used.
In GANs~\citep{goodfellow2014generative}, latent expressions are acquired by training two models, a Generator and a Discriminator, alternately.
The Generator creates data variables $\boldsymbol{\theta}_{\mathrm{model}}$ from latent variables $\boldsymbol{z}$.
The Discriminator estimates whether given variables are a sample from the data set $\boldsymbol{\theta}_{\mathrm{non \mathchar`- collision}}$ or a generated sample $\boldsymbol{\theta}_{\mathrm{model}}$ calculated from $\boldsymbol{z}$, which is uniformly sampled from the latent space within [0,1].
Since the latent space is a convex space and the boundaries of the latent space can be arbitrarily determined in advance, any point of a line segment connecting any point is in that latent space, within a domain of definition.
Furthermore, it is possible to give conditions to the models by introducing a $Condition$ variable $\boldsymbol{c}$~\citep{mehdi2014conditional}.

Figure~\ref{fig:network} shows the concept of the proposed network model.
Through the Generator, the mapping from the latent space to collision-free joint space is obtained.
The Discriminator identifies the joint angles, generated by the Generator $\boldsymbol{\theta}_{\mathrm{model}}$, and the actual sampled joint angles, $\boldsymbol{\theta}_{\mathrm{non \mathchar`- collision}}$.
In condition $\boldsymbol{c}$, the obstacle information is given as a binary mask showing the location of the obstacle.
This condition $\boldsymbol{c}$ is connected to the Generator and the Discriminator, so that when the given obstacle mask changes, the correspondence from the latent space to joint space changes.
In other words, our method does not need to prepare a different network for each obstacle, and only one cGANs can support multiple obstacle environments.

The loss function, $\mathcal{L}_{\mathrm{GAN}}$, for training cGANs is shown in  equation~(\ref{eq:loss_gan}).
\begin{equation}
    \label{eq:loss_gan}
    \begin{split}
    \mathcal{L}_{\mathrm{GAN}}(D, G) = &
        \mathbb{E}_{ 
            \boldsymbol{c} \sim p_{\mathrm{obstacles}} \left( \boldsymbol{c} \right), \,
            \boldsymbol{\theta} \sim p_{\mathrm{non \mathchar`- collision}} \left( \boldsymbol{\theta} | \boldsymbol{c} \right)
        } \left[ 
            \log D \left( \boldsymbol{\theta} , \boldsymbol{c} \right) 
        \right] \\
         & + \mathbb{E}_{
            \boldsymbol{c} \sim p_{\mathrm{obstacles}} \left( \boldsymbol{c} \right), \,
            \boldsymbol{z} \sim p_{\boldsymbol{z}} \left(\boldsymbol { z }\right)
        } \left[
            \log \left( 
                1 - D \left( G \left( 
                         \boldsymbol{z} , \boldsymbol{c} 
                    \right) , \boldsymbol{c} 
                \right) 
            \right) 
        \right] 
    \end{split}
\end{equation}
\noindent
Where $p_{\mathrm{obstacles}}(\boldsymbol{c})$ is the distribution of obstacles positions and
$p_{\mathrm{non \mathchar`- collision}}(\boldsymbol{\theta}|\boldsymbol{c})$ is the distribution of non-collision joint angles which the Generator should aim to generate.
$p_{\boldsymbol{z}}(\boldsymbol{z})$ is the uniform distribution in the latent space.

\subsection{$\mathcal{L}_{\mathrm{collision}}$: Adaptation to Multiple Obstacle Conditions}
In this section, we describe how to adapt to various obstacle conditions.
Even though collision-free mapping from the latent space to joint space is trained by equation~(\ref{eq:loss_gan}), the network cannot learn well since the number of non-collision data points is much smaller than those with collisions.
As the obstacles become more diverse, there is a risk of mistaking collision points for non-collision points and vice versa.

It is therefore necessary to train with the collision joints explicitly incorporated within the equation.
The loss function, $\mathcal{L}_{\mathrm{collision}}$, shown in equation~(\ref{eq:loss_col}) is introduced in order to provide the data of the collision joints to the Discriminator.
\begin{equation}
    \label{eq:loss_col}
    \mathcal{L}_{\mathrm{collision}}(D) = 
        \mathbb{E}_{ 
            \boldsymbol{c} \sim p_{\mathrm{obstacles}} \left( \boldsymbol{c} \right), \,
            \boldsymbol{\theta} \sim p_{\mathrm{collision}} \left( \boldsymbol{\theta} | \boldsymbol{c} \right)
        } \left[
            \log \left( 
                1 - D \left( \boldsymbol{\theta} , \boldsymbol{c} 
                      \right) 
                 \right) 
        \right] 
\end{equation}
\noindent
Where $p_{\mathrm{collision}}(\boldsymbol{\theta} | \boldsymbol{c} )$ is the distribution of joint angles that collide with obstacles, which the Generator should thus refrain from generating.
The Discriminator is trained to output $0$ for collision joints and $1$ for collision-free joints for each obstacle. 
Furthermore, the Generator is trained to acquire a distribution to make the Discriminator output $1$, as we are trying to obtain a distribution for collision-free space.

\subsection{$\mathcal{L}_{\mathrm{identity}}$: Specifying the Map from the Latent Space to Joint Space}
We will describe in this section how to map from the latent space to joint space, such that arbitrary planned paths in the latent space are smooth in joint space for robot arms.
In equation~(\ref{eq:loss_gan}), the path planned in the latent space is mapped from each point in the latent space to joint space, but it is not certain whether the path planned in the latent space can be realized by the robot in joint space.
The mapping from the latent space to joint space has to be continuous without ``twists'',``distortions'', and rapid changes.
In order to achieve this, the following two things are performed:
\begin{itemize}
    \item The number of dimensions for latent variables is matched to the number of robot joints; each latent variable is mapped to represent each joint, and the normalized ranges of latent variables and joint angles are aligned.
    \item The Generator is trained to output $\boldsymbol{\theta}$ when the latent variables $\boldsymbol{z}=\boldsymbol{\theta}$ are given as input of the Generator.
    In other words, a certain distance in the latent space is almost the same distance in joint space.
\end{itemize}
However, since the acquired map may be distorted in order to avoid collisions, these constraints are not added to the joint that collides with the obstacles.
The loss function, $\mathcal{L}_{\mathrm{identity}}$, for training cGANs is shown in equation~(\ref{eq:loss_im}).
\begin{equation}
    \label{eq:loss_im}
    \mathcal{L}_{\mathrm{identity}}(G) = 
        \mathbb{E}_{ 
            \boldsymbol{c} \sim p_{\mathrm{obstacles}} \left( \boldsymbol{c} \right), \,
            \boldsymbol{\theta} \sim p_{\mathrm{non \mathchar`- collision}} \left( \boldsymbol{\theta} | \boldsymbol{c}  \right)
        } \left[
            \| 
                G \left( \boldsymbol{z} = \boldsymbol{\theta} , \boldsymbol{c} 
                \right) - \boldsymbol{\theta}
            \|_{2}^{2}
        \right] 
\end{equation}


\section{Experiment}
\label{sec:experiment}
The purpose of the experiment is to confirm three factors, \rnum{1}~)~\textbf{\emph{Adaptation}}, \rnum{2}~)~\textbf{\emph{Customizability}}, and \rnum{3}~)~\textbf{\emph{Scalability}}, by the proposed mapping function from the latent space to joint space acquired by training cGANs. 
We performed an experiment using a simulation of a two-link arm robot as the simplest example to analyze the results in detail.

\subsection{Robot Setup}
\begin{figure}[tb]
    \centering
    \includegraphics[width=0.50\linewidth]{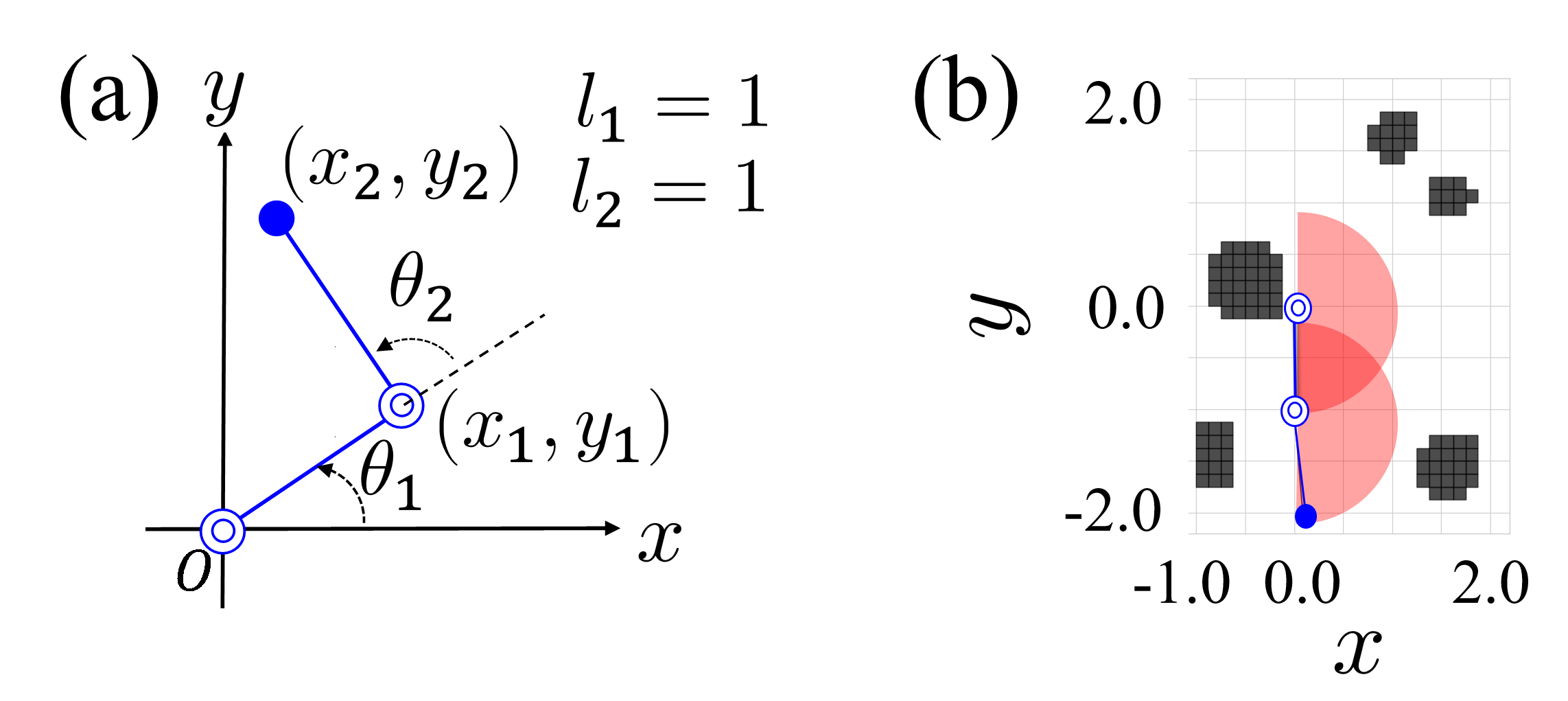}
    \caption{
		(a) A simulator of a two-link robot arm with parameters, (b) an example of an environment with obstacles.
		No obstacles are placed in the red area to avoid narrowing the robot's workspace.
		}
    \label{fig:experiment_robot}
    \vspace{-4mm}
\end{figure}

To perform our experiments, we use a simulation of a two-link robot arm (See Figure~\ref{fig:experiment_robot} (a)).
The joint angles of the first and second axes are $\theta_1$ and $\theta_2$, respectively, and the length of each link of the arm is $1.0$.
The range of each joint angle is $\theta_1 \in [-90^\circ, 90^\circ]$ and $\theta_2 \in [5^\circ, 150^\circ]$.
Considering the $xy$ plane with the first joint of the robot as the origin, the obstacles are arranged in a grid pattern in the range of $-1.0 \leq x \leq 2.0$, $-2.0 \leq y \leq 2.0$ (See Figure~\ref{fig:experiment_robot} (b)).

\subsection{Data Collection}
We will describe the details of how the data was collected in the following subsections. 

\subsubsection{Robot movement}
The training data of cGANs consists of two joint angles $\boldsymbol{\theta}=(\theta_1, \theta_2)$ that represents the posture of a two-link robot arm.
The latent variables $\boldsymbol{z}$ are also two-dimensional because the robot has 2-DoF, and a uniform distribution in the range $[0, 1]$ is used.
The step sizes for sampling the joint angles are both set to $5^\circ$.
$37\times30=1110$ data points for each obstacle condition is created.
Min-Max scaling was applied to each of $\theta_1, \theta_2$, and both were normalized to the range of $[0, 1]$.

\subsubsection{Obstacle Position}
The position of the obstacle is given as a binary mask as Condition $\boldsymbol{c}$, and a position occupied by the obstacle is represented as $1$, and an unoccupied position is represented by $0$.
The Condition $\boldsymbol{c}$ is a two-dimensional array of $32\times24$.
Obstacles are placed in specific units by randomly deciding the position and size of a circle or rectangle with a minimum size of 0.15 in length and 0.15 in width. 
However, there is a constraint that prevents obstacles from being placed in places that greatly obstruct the movement of the robot, in order to prevent the robot's workspace from being too narrow (See Fig.~\ref{fig:experiment_robot}(b)).
For collision detection, we checked whether the four sides of each obstacle intersects with either of the two links of the robot arm.

\subsubsection{Dataset}
There is one dataset without obstacle conditions and 1,000 datasets with obstacle conditions.
5-fold cross-validation is performed using $800$ for training and $200$ for evaluation.

\subsection{Network Design}
The architecture of our network model is composed of a Generator and a Discriminator with fully connected layers, and each network includes a two-dimensional convolutional layer as a feature extraction unit for Conditions (See Figure~\ref{fig:network}).
For learning stabilization, batch normalization~\citep{batch_norm}, spectral normalization~\citep{sngan}, and feature matching~\citep{feature_match} were used.
Our network model is implemented with Chainer~\citep{chainer_learningsys2015} as deep learning library and detailed parameters are shown in the supplemental material.
Training is conducted on a machine equipped with Intel Core i9-7900X CPU and GeForce RTX 2080, resulting in about 1 to 1.5 days of training time.

\section{Results}
\label{sec:result}
\subsection{Mapping from Latent Space to Joint Space}
\begin{figure}[tb]
    \centering
    \includegraphics[width=0.95\textwidth]{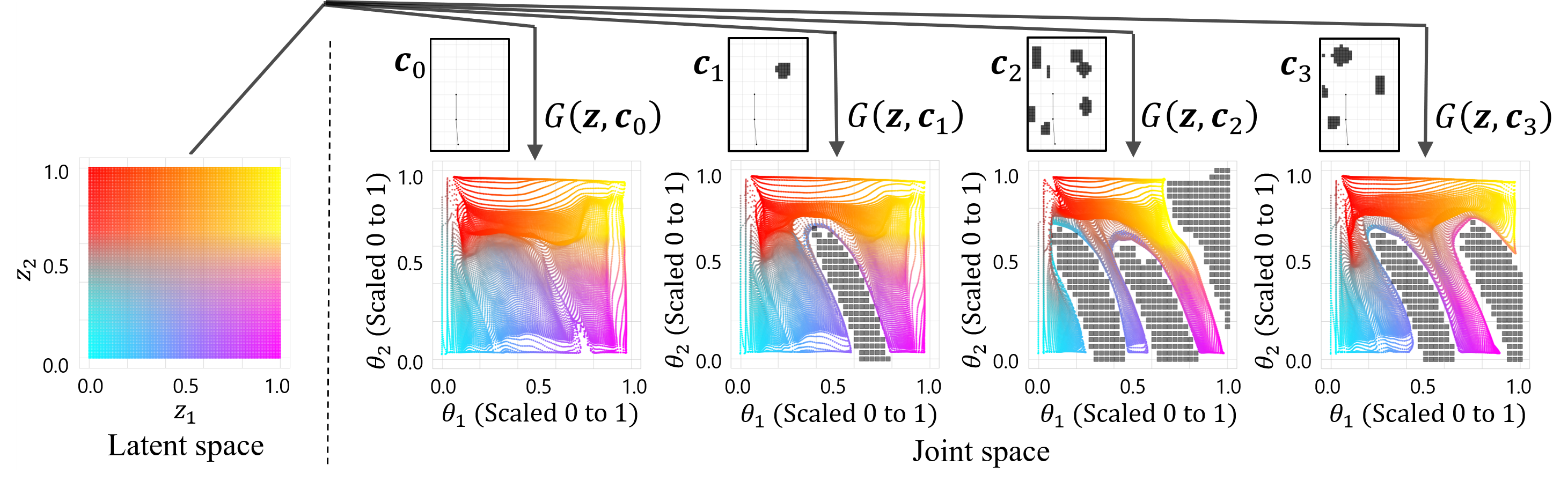}
    \caption{
    Acquired mappings from latent space to joint space with different obstacle positions, $\boldsymbol{c}_{0} \sim \boldsymbol{c}_{3}$.
    Areas in the latent space that correspond with areas in joint space are shown in the same colors.
    The gray area in the joint space represents the joint angles that collide with obstacles.
    }
    \label{fig:result_mapping}
    \vspace*{-4mm}
\end{figure}

We first show examples of the mapping from the latent space to joint angle space acquired by learning (Figure~\ref{fig:result_mapping}).
Areas in the latent space that correspond with areas in joint space are shown in the same colors.
We can observe that approximately $z_1$ and $\theta_1$ and $z_2$ and $\theta_2$ have a corresponding relationship between the latent space and obstacle-free joint space (See the latent space and joint space without obstacles, condition $\boldsymbol{c}_0$, in Figure~\ref{fig:result_mapping}).
It can also be seen that the mapping from latent space to joint space corresponds to each other continuously, without torsions or distortions.
Additionally, the mapping was generated while avoiding the obstacles' areas according to the given conditions. 

\subsection{Collision-free Path Planning}
\begin{figure}[tb]
    \centering
    \includegraphics[width=0.80\textwidth]{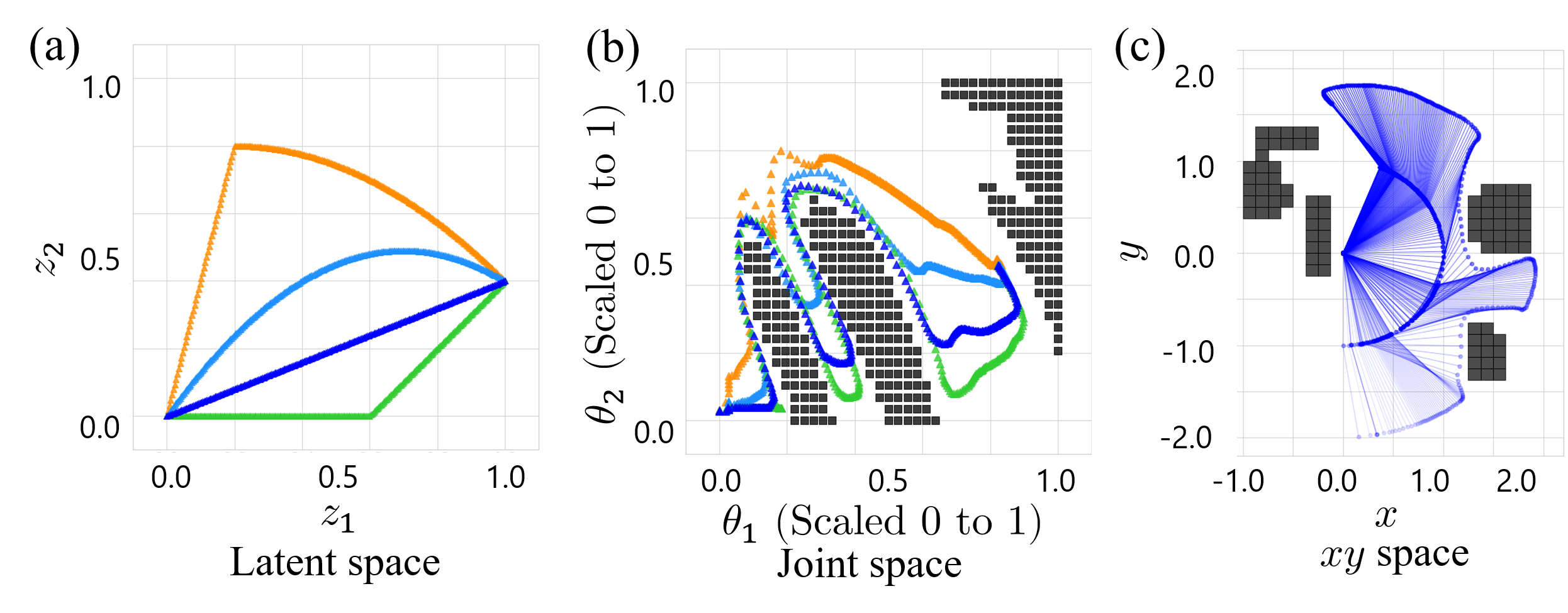}
    \caption{
    Examples of path planning in latent space for untrained conditions.
    (a) the latent space, (b) the joint space, and (c) the $xy$ space.
    The paths of the same color in the latent space and the joint space correspond to each other.
    The blue planned path is also shown in $xy$ space.
    }
    \label{fig:trajectory_success}
\end{figure}

Here, we show examples of collision-free path planning in the latent space.
Figure~\ref{fig:trajectory_success} shows the result of mapping planned paths in the latent space to joint space when an untrained condition is given (See more examples in the attached video).
Any planned paths in the latent space is collision-free in joint space.
The path planning sometimes failed because it touched the surface of the obstacle at the collision/non-collision boundary.
In the next section, we will describe how accurate it is.

\subsection{Quantitative Evaluation of Acquired Mapping}
\begin{figure}[tb]
    \centering
    \includegraphics[width=0.50\linewidth]{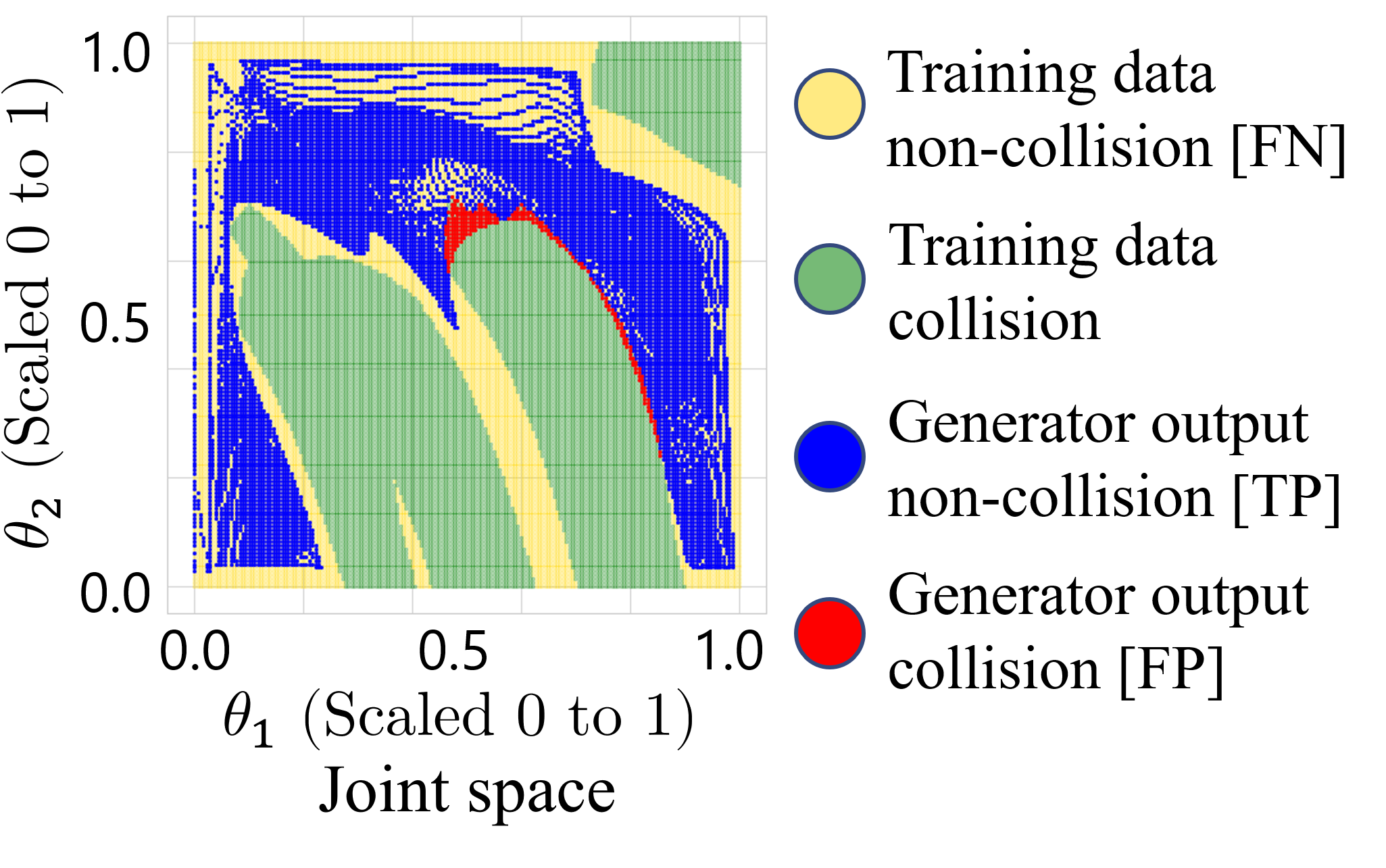}
    \caption{
        An example of the metric calculation process.
        The joint angle space is divided into grids.
        Each point is marked Positive/Negative which indicates whether the point is generated by $G$ or not and True/False, which indicates whether the output is correct or not.
        }
    \label{fig:metrics_example}
\end{figure}

We evaluate the accuracy of the mapping acquired from the Generator to  verify \rnum{1}~)~\textbf{\emph{Adaptation}}.
Both joint and the latent spaces are divided at regular intervals into a grid, by $\Delta \theta$ and similar values from the latent space.
Joint angles are generated by the Generator by sampling one point from each grid point in the latent space.
The generated joint angles are assigned to the corresponding grid points in joint space.
Of all the grid points in the joint space, we counted the following numbers:
\begin{itemize}
    \item the grid points that avoid collisions are True Positive (TP),
    \item the grid points that do not avoid collisions but are not generated is False Negative (FN),
    \item and the grid points that do not avoid collisions but are generated, are False Positive (FP).
\end{itemize}
Using these numbers, the accuracy was evaluated by following two metrics: 1) the overlapping area between the joint space area that avoided collisions and the area generated by the Generator (calculated by equation~(\ref{eq:iou})) using the Intersection over Union (IoU), 2) the proportion of the joint angles generated by the Generator that actually avoided collisions (calculated by equation~(\ref{eq:precision}) as precision) (See Figure~\ref{fig:metrics_example} as an example of the process for calculating metrics.)
\vspace{-1.0mm}
\begin{eqnarray}
    \mathrm{IoU} \left( G, \, \boldsymbol{c} \right) &=&
    \mathrm{TP}/(\mathrm{TP}+\mathrm{FP}+\mathrm{FN})
    \label{eq:iou}
    \\
    \mathrm{Precision} \left( G, \, \boldsymbol{c} \right) &=&
    \mathrm{TP}/(\mathrm{TP}+\mathrm{FP})
    \label{eq:precision}
\end{eqnarray}
\vspace{-7.0mm}

Table~\ref{table:result_cv_mean} shows the average value of IoU and the precision in 5-fold cross validation.
The results depend on the size of the grid points, but we confirmed that the results converged when $\Delta \theta = 1^\circ$ and the total number of samples $N$ was $26,426$.
The average value of IoU between trained and untrained conditions is similar to the distribution of joint angles that avoid collisions.
The average precision in both trained and untrained conditions is high.
We performed 5-fold cross-validation and select the best IoU averages, shown in Figure~\ref{fig:metrics_hist} as a histogram.
The relative frequency on the vertical axis is divided by the number of data points (801 training and 200 test data points).
The IoU is around $0.6$ in many conditions for both the training data and the test data, but the histogram spreads to the left in the test data, which means the value of IoU is worse on some Conditions in the test data.
This result suggests that our single trained model can \textbf{adapt} to multiple obstacle conditions including untrained ones.

\begin{table}[tb]
    \begin{minipage}{0.48\linewidth}
        \tblcaption{Statistics of accuracy of the mapping by the Generator in the 5-fold cross-validation.}
        \label{table:result_cv_mean} 
        \centering
        \begingroup
            \scalefont{0.9}
            \begin{center}
            \scalebox{1.00}{
                \begin{tabular}{c c c c c}
                \hline 
                \multicolumn{1}{c}{} &
                \multicolumn{4}{c}{IoU} \\
                \cline{1-5}
                \multicolumn{1}{c}{} & mean & std & min & max\\
                \hline
                Train & \textbf{0.493} & 0.022 & 0.407 & 0.549 \\
                Test & \textbf{0.491} & 0.024 & 0.404 & 0.543 \\
                \hline \hline
                \multicolumn{1}{c}{} & \multicolumn{4}{c}{Precision}\\
                \cline{1-5}
                \multicolumn{1}{c}{} & mean & std & min & max\\
                \hline
                Train & \textbf{0.990} & 0.006 & 0.947 & 0.997 \\
                Test  & \textbf{0.986} & 0.013 & 0.904 & 0.997 \\
                \hline 
                \end{tabular}
                }
            \end{center}
    \endgroup

    \end{minipage}
    \begin{minipage}[c]{0.48\linewidth}
        \centering
        \includegraphics[width=0.9\linewidth]{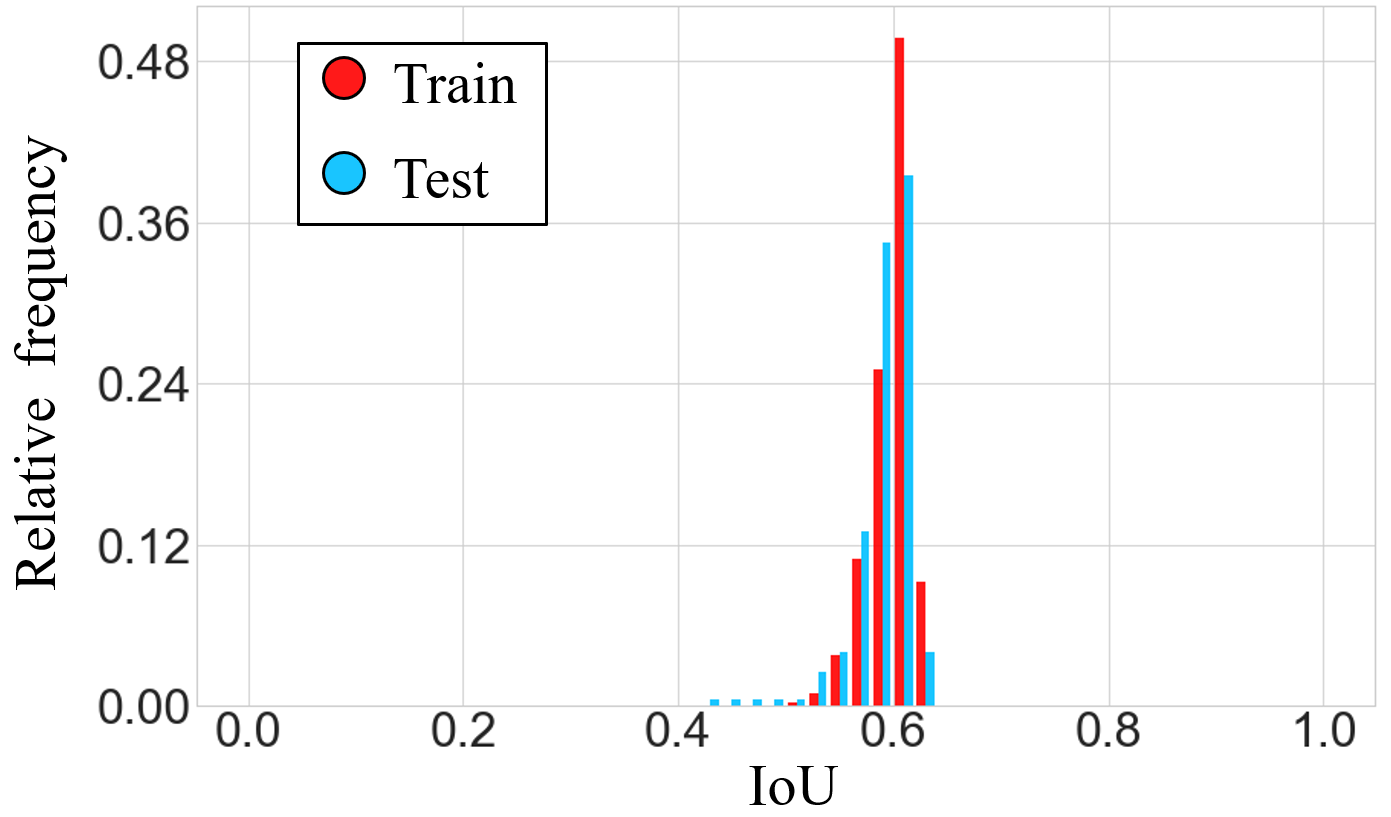}
        \figcaption{
            A histogram of the best IoU averages from 5-fold cross validation.
            }
        \label{fig:metrics_hist}
    \end{minipage}
\end{table}

\subsection{Path Selection in the Latent Space}
\begin{figure}[tb]
    \centering
    \includegraphics[width=0.80\linewidth]{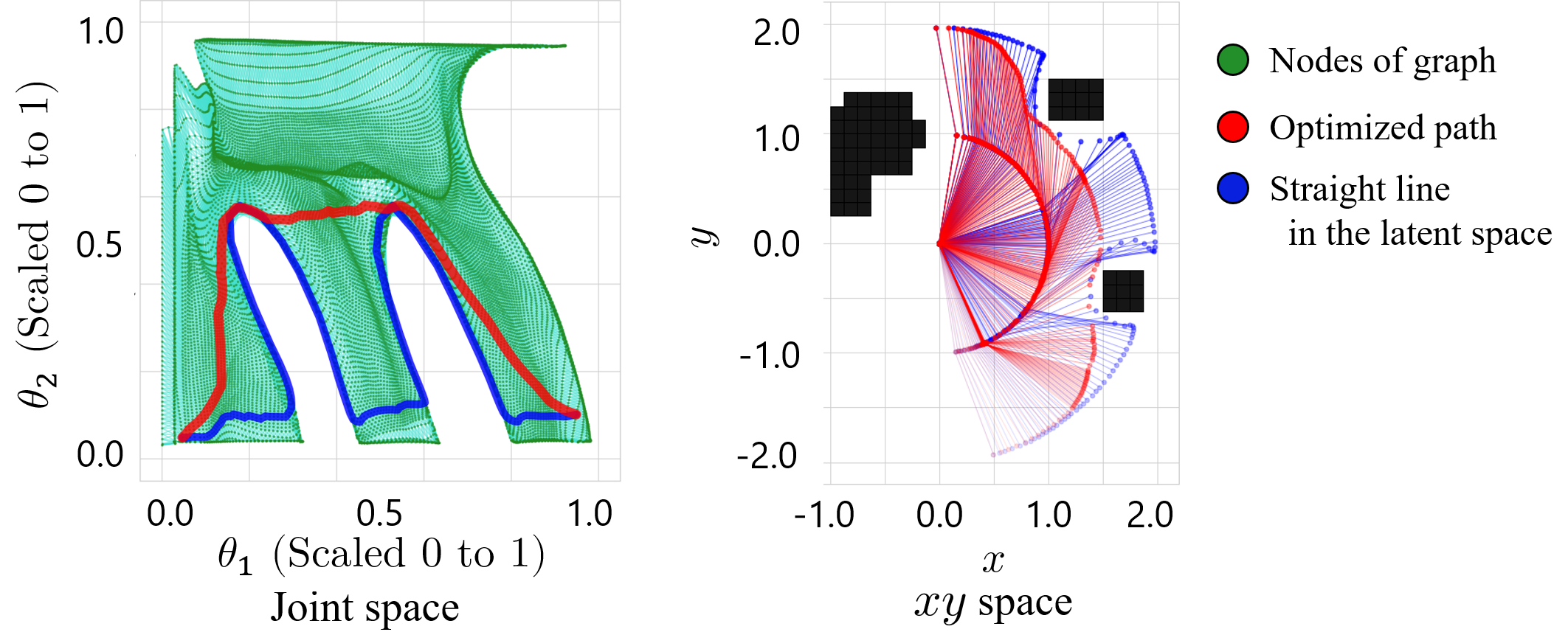}
    \caption{
        One of the examples of the optimized path.
        The red path is optimized for \textbf{the shortest traveling distance in joint space} by $A^{\ast}$, and the blue path is a straight line in the latent space.
        }
    \label{fig:Astar}
    \vspace{-4mm}
\end{figure}

Here, we verify \rnum{2}~)~\textbf{\emph{Customizability}} and \rnum{3}~)~\textbf{\emph{Scalability}}.
The proposed method can generate multiple (in-finite) paths.
No matter what kind of path is planned in the latent space, the result is mapped to the corresponding collision-free path in joint space.
The method for determining paths in the latent space is not limited to just connecting the start and the goal linearly, but can be any path/trajectory planner; this makes our method highly \textbf{\emph{customizable}} since we can use anything we want.
One idea is to use graph optimization for the shortest traveling distance in joint space.
Figure~\ref{fig:Astar} shows the result of constructing a graph from neighboring points in the latent space and performing $A^{\ast}$ optimization in joint space.
The heuristic function used is the Euclidean distance.
Nodes of the graph are uniformly sampled in $z_1, z_2$ and a total of $N=128 \times 128 = 16,384$.

\begin{figure}[tb]
    \centering
    \includegraphics[width=0.80\linewidth]{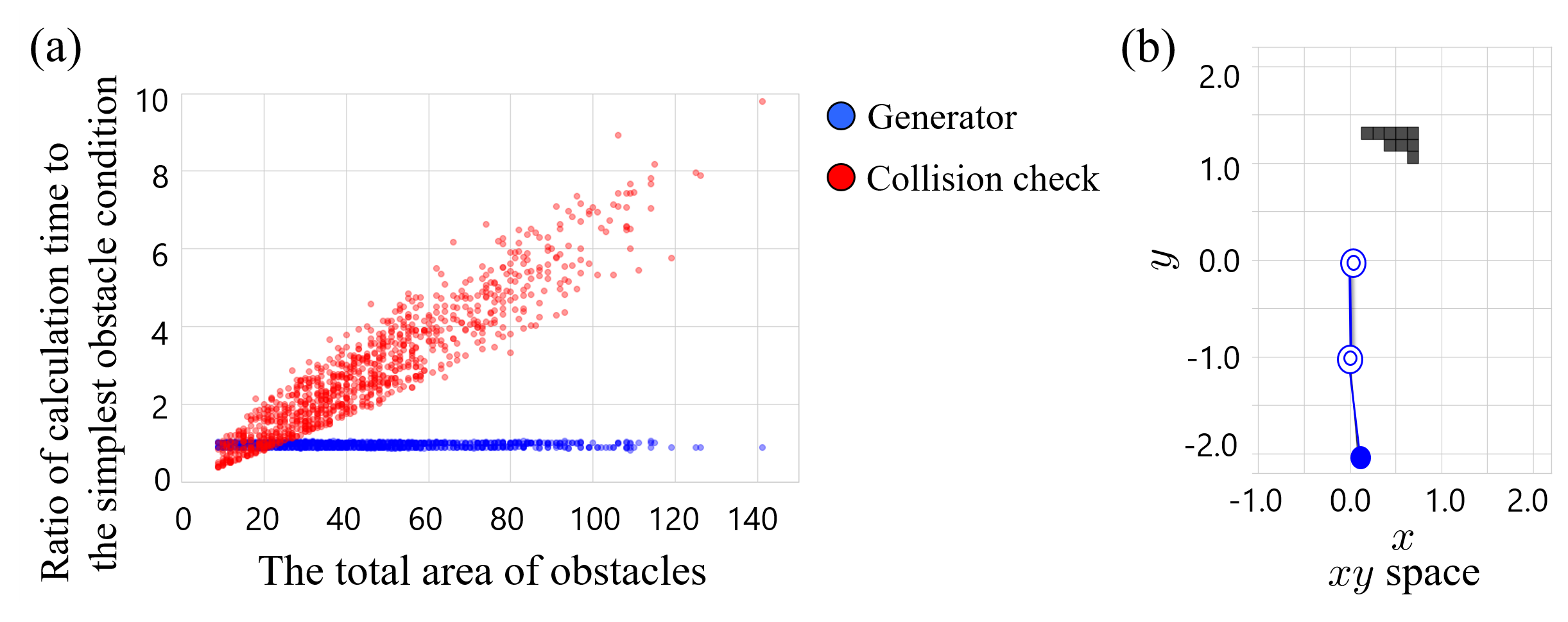}
    \caption{
        (a) The relationship between the number of obstacles and the calculation time.
        In order to reduce the effect of program performance, we used the ratio of CPU time to computation time for each label under the simplest obstacle condition, (b).
        }
    \label{fig:cpu_time}
\end{figure}

Figure~\ref{fig:cpu_time} shows the relationship between the total area of obstacles and CPU calculation time for the number of nodes described above.
The computational cost for collision checks increases as the number of obstacles increases and as the shapes become complicated, but the computational cost of the inference by the Generator does not depend on the obstacles after the learning phase since collision checks are not required, making the computations \textbf{\emph{scalable}} to the number of obstacles.
The inference by the Generator for $16,384$ latent variables is computed by the GPU in 0.119 seconds.

\section{Discussion about Application to High-DoF Robots}
\label{sec:Discussion}
The proposed method can be used for robots with higher DoF.
In this research, we conducted experiments using a two-link robot arm accompanying 2-dimensional latent space.
The proposed network model can be applied if the dimension of the latent space is increased according to the DoF of the robot.
The condition variables can be given in 3D, so 3D convolution can be applied.

One of the main challenges is that the training dataset size will increase as the number of dimensions increases.
In this study, the robot was trained in the entire motion range of the robot, but by targeting the work and motion range with a focus on specific tasks, the data size can be reduced.
\section{Conclusion}
\label{sec:conclusion}
In this research, the robot's collision-free joint angles are expressed as the latent space of cGANs, and collision-free paths are obtained by mapping the path planned in the latent space to joint space.
It was confirmed that \rnum{1}~)~\textbf{\emph{Adaptation}}; a single trained model could handle multiple unknown obstacle conditions, \rnum{2}~)~\textbf{\emph{Customizability}}; any path can be planned in the latent space, and \rnum{3}~)~\textbf{\emph{Scalability}}; computational cost of path planning does not depend on the obstacles.
As for future prospects, we are considering the application of this method to higher DoF robots as discuss in section~\ref{sec:Discussion}.
\clearpage

\acknowledgments{
    The authors would like to thank Wilson Ko for helping writting and proofreading.
    Hiroki Mori would like to thank all colleagues in ETIS lab in the University of Cergy-Pontoise, especially Prof. Mathias Quoy, Prof. Philippe Gaussier and Assoc. Prof. Alexandre Pitti because he came up with the basic idea of this article and discussed a preliminary result with them when he joined the lab in 2016.
}

\bibliography{bibliography}
\end{document}